\documentclass{article}

\usepackage{iclr2027_conference,times}

\usepackage[utf8]{inputenc}
\usepackage[T1]{fontenc}
\usepackage{hyperref}
\usepackage{url}
\usepackage{xurl}
\usepackage{microtype}
\usepackage{graphicx}
\usepackage{subcaption}
\usepackage{booktabs}
\usepackage{multirow}
\usepackage{array}
\usepackage{arydshln}
\usepackage{dashrule}
\usepackage{makecell}
\usepackage{amsmath}
\usepackage{amssymb}
\usepackage{amsfonts}
\usepackage{mathtools}
\usepackage{amsthm}
\usepackage{bm}
\usepackage{xcolor}
\usepackage[capitalize,noabbrev]{cleveref}

\theoremstyle{plain}

\theoremstyle{definition}

\theoremstyle{remark}

\usepackage{float}

\usepackage{pgfplots}
\pgfplotsset{compat=1.18}
\usepgfplotslibrary{groupplots}

\title{PrefillShare: A Shared Prefill Module for KV\\
Reuse in Multi-LLM Disaggregated Serving}

\author{\parbox[t]{\dimexpr\textwidth-2\tabcolsep\relax}{%
\centering\bfseries
Sunghyeon Woo\textsuperscript{1}\thanks{Equal contribution.}\quad
Hoseung Kim\textsuperscript{1}\footnotemark[1]\quad
Sunghwan Shim\textsuperscript{1}\quad
Minjung Jo\textsuperscript{2}\thanks{Work done while at NAVER Cloud.}\\[3pt]
Hyunjoon Jeong\textsuperscript{1}\quad
Jeongtae Lee\textsuperscript{1}\quad
Joonghoon Kim\textsuperscript{1}\quad
Sungjae Lee\textsuperscript{1}\\[3pt]
Jiwon Ryu\textsuperscript{1}\quad
Baeseong Park\textsuperscript{3}\footnotemark[2]\quad
Se Jung Kwon\textsuperscript{3}\footnotemark[2]\quad
Dongsoo Lee\textsuperscript{3}\footnotemark[2]\\[7pt]
\normalfont
\textsuperscript{1}\,NAVER Cloud\qquad
\textsuperscript{2}\,Toss\qquad
\textsuperscript{3}\,A2SYS\\[3pt]
\texttt{\{sunghyeon.woo1, hoseung.kim\}@navercorp.com}
}}

\iclrfinalcopy
\hypersetup{
  pdftitle={PrefillShare: A Shared Prefill Module for KV Reuse in Multi-LLM Disaggregated Serving},
  pdfauthor={Sunghyeon Woo, Hoseung Kim, Sunghwan Shim, Minjung Jo, Hyunjoon Jeong, Jeongtae Lee, Joonghoon Kim, Sungjae Lee, Jiwon Ryu, Baeseong Park, Se Jung Kwon, Dongsoo Lee}
}
\begin{document}
\maketitle
\lhead{}

\begin{abstract}
 Multi-agent systems increasingly orchestrate multiple specialized language models to solve complex real-world problems, often invoking them over a shared context. This execution pattern repeatedly processes the same prompt prefix across models. Consequently, each model redundantly executes the prefill stage and maintains its own key–value (KV) cache, increasing aggregate prefill load and worsening tail latency by intensifying prefill–decode interference in existing LLM serving stacks. Disaggregated serving reduces such interference by placing prefill and decode on separate GPUs, but disaggregation does not fundamentally eliminate inter-model redundancy in computation and KV storage for the same prompt. To address this issue, we propose PrefillShare, a novel algorithm that enables sharing the prefill stage across multiple fine-tuned models in a disaggregated setting. PrefillShare factorizes the model into prefill and decode modules, freezes the prefill module, and fine-tunes only the decode module. The shared-prefill design allows multiple task-specific models to share a prefill module and the KV cache generated for the same prompt when prefix overlap is high, while shared prefill workers pool prefill capacity across models and mitigate load imbalance even when prefix overlap is limited. We further implement a routing and cache-handoff mechanism in a vLLM-based disaggregated system to support shared-prefill execution across heterogeneous specialized decoders. PrefillShare not only matches full fine-tuning accuracy on a broad range of tasks and models, but also delivers 4.5$\times$ lower p95 latency and 2.0$\times$ higher throughput in multi-model agent workloads. 
\end{abstract}

\section{Introduction}
Large language models (LLMs) have demonstrated strong performance across reasoning, code generation, and knowledge-intensive question answering~\citep{chowdhery2023palm, team2023gemini}, motivating agentic workflows in which models use tools, memory, and external resources~\citep{schick2023toolformer, shinn2023reflexion, yao2022react}. Recent systems extend this paradigm to multi-model workflows, where specialized models collaborate by leveraging distinct fine-tuned capabilities~\citep{hong2024metagpt, su2025toolorchestra, wu2024autogen}. While these workflows improve modularity and task performance, they introduce substantial inference inefficiencies when multiple models operate on a shared context. Modern serving systems rely on KV caches to avoid recomputing attention over previous tokens in decoder-only Transformers~\citep{kwon2023efficient, zheng2024sglang}, but KV caches are tied to model parameters and cannot be reused across differently fine-tuned models. As a result, each model must independently prefill the same context and maintain its own KV cache, increasing time-to-first-token (TTFT) and tail latency due to prefill/decode contention while also inflating memory usage. Disaggregated serving mitigates this contention by separating the compute-bound prefill phase from the memory-bound decoding phase~\citep{patel2024splitwise,zhong2024distserve}; however, it does not eliminate cross-model redundancy, leaving identical prompts to incur duplicated prefill computation and KV-cache across specialized models.

\begin{figure*}[t]
    \centering
    \includegraphics[width=1.0\textwidth]{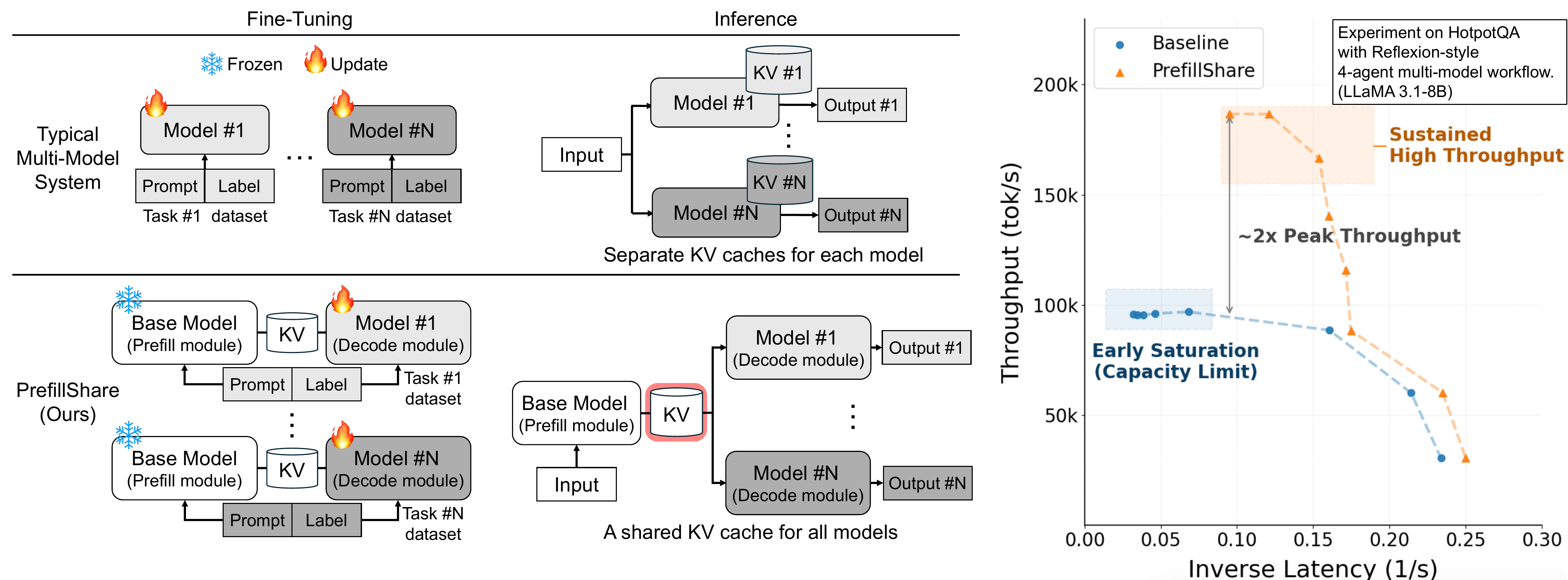}
    \caption{
    \textbf{Comparison between conventional multi-model serving and PrefillShare.}
    \textbf{(Left)} Typical multi-model systems fine-tune and serve each task-specific model independently, causing redundant prefill computation and KV caches for identical prompts. PrefillShare partitions a decoder-only Transformer into a shared prefill module and task-specific decode modules, allowing multiple specialized decoders to reuse the same prompt KV cache.
    \textbf{(Right)} In disaggregated serving, redundant prefill and per-model KV-cache growth increase computation and trigger cache evictions under high load. By sharing prompt KV caches across models, PrefillShare improves prefix-cache reuse, delays cache saturation, and achieves up to $2\times$ higher throughput.
    }
    \label{fig:architecture}
\end{figure*}

Recent works have begun to address these inefficiencies through cross-model KV reuse. KVCOMM~\citep{ye2025kvcomm} reuses KV cache across overlapping contexts by aligning cache offsets with an online anchor pool, while DroidSpeak~\citep{liu2024droidspeak} enables partial KV sharing by reusing states from non-sensitive components under specific architectural assumptions. Although these methods show the promise of cross-model reuse, they typically require additional coordination and may not generalize robustly without explicit training for KV-cache sharing. ICaRus~\citep{woo2026icarus} takes a different approach by decomposing a decoder-only Transformer into a frozen logical encoder and a trainable logical decoder, enabling identical KV reuse across fine-tuned models for the same prompt. However, since its decoder consumes only encoder-generated KV caches, the encoder must still run during decoding, adding compute overhead and limiting scalability under high concurrency.

In this work, we propose PrefillShare, a shared-prefill disaggregated serving algorithm for multi-model agent workloads. As illustrated in Fig.~\ref{fig:architecture}, PrefillShare partitions a decoder-only Transformer into a shared prefill module and task-specific decode modules: the prefill module processes the shared context once to produce a reusable KV cache, while specialized decoders generate outputs conditioned on this cache. To enable robust shared-prefill execution, PrefillShare employs cache-conditioned fine-tuning, which freezes the prefill module and trains each task-specific decoder to consume the base model’s KV cache. We further implement efficient routing and cache handoff in a vLLM-based disaggregated system to preserve shared-cache locality across heterogeneous model switches. PrefillShare not only eliminates redundant prefill computation and duplicated KV storage when prefix overlap is high, but also pools prefill capacity across models to mitigate load imbalance when prefix overlap is limited. As a result, PrefillShare preserves the accuracy of full fine-tuning across diverse tasks and models while substantially reducing tail latency and improving throughput in multi-model serving workloads.

In summary, we make the following contributions:
\begin{itemize}
  \item \textbf{Cross-model prefill and KV-cache sharing.}
  We propose PrefillShare, a novel shared-prefill serving algorithm that allows multiple task-specific models to reuse the same prefill computation and KV cache for identical prompts in disaggregated serving, eliminating cross-model redundancy in both computation and memory.

  \item \textbf{Cache-conditioned fine-tuning for robust sharing.}
  We introduce a disaggregation-aware fine-tuning procedure that freezes the shared prefill module and trains only task-specific decode modules conditioned on the shared KV cache, aligning training with shared-prefill inference while matching full fine-tuning accuracy.

  \item \textbf{Efficient routing and cache handoff.}
  We implement a routing and cache-handoff mechanism in a vLLM-based heterogeneous disaggregated serving system for different fine-tuned models, making shared-prefill inference practical and achieving up to \textbf{4.5$\times$} lower p95 latency and \textbf{2.0$\times$} higher throughput on multi-model agent workloads. 
\end{itemize}

\section{Background}
\label{sec:background}

\subsection{Autoregressive Inference and KV Caching}

Let $X = \{x_1, \dots, x_n\}$ denote an input prompt of length $n$, and let $Y = \{y_1, \dots, y_T\}$ denote the generated sequence of length $T$. An LLM parameterized by $\theta$ models the conditional likelihood as
\begin{equation}
    P(Y | X) = \prod_{t=1}^{T} P(y_t | X, y_{<t}; \theta),
\end{equation}
where $y_{<t}$ denotes previously generated tokens.

Decoder-only Transformers avoid redundant attention computation by maintaining a KV cache. We write the Transformer forward pass as
\begin{equation}
    (y, \Delta \mathcal{C}) = \mathcal{F}_{\theta}(x, \mathcal{C}_{\text{past}}),
\end{equation}
where $x$ is the current input, $\mathcal{C}_{\text{past}}$ is the existing cache, $y$ is the next token, and $\Delta \mathcal{C}$ is the newly generated KV state. In the prefill phase, the model processes the entire prompt in parallel:
\begin{equation}
    (y_1, \mathcal{C}_n) = \mathcal{F}_{\theta}(X, \emptyset).
    \label{eq:prefill}
\end{equation}
It then enters the decode phase, generating tokens autoregressively as
\begin{equation}
    (y_t, \Delta \mathcal{C}_t) =
    \mathcal{F}_{\theta}(y_{t-1}, \mathcal{C}_{n+t-2}), \quad
    \mathcal{C}_{n+t-1} = [\mathcal{C}_{n+t-2}; \Delta \mathcal{C}_t].
    \label{eq:decode}
\end{equation}
Since the cache grows linearly with sequence length and must be repeatedly accessed during decoding, the decode phase is typically memory-bound.

\subsection{Multi-Model Inference and Prefix Caching}

Agentic workflows increasingly deploy multiple specialized models $\{\mathcal{M}_1, \dots, \mathcal{M}_N\}$ to solve complex tasks. In such workflows, different models often process the same shared context $X$, such as a long conversation history, a common system prompt, or a shared document, while producing task-specific outputs. This creates repeated prefill computation over identical or highly overlapping contexts across models.

A standard way to avoid repeated prefill within a single model is prefix caching~\citep{kwon2023efficient, zheng2024sglang}. For an incoming request with a prefix $P$, the serving system checks whether the corresponding prefill cache $\mathcal{C}_{|P|}$, produced by Eq.~\eqref{eq:prefill}, already exists in the cache pool. If a match is found, the cached KV blocks are reused and the corresponding prefill computation is skipped, turning repeated prefill into a lightweight cache lookup.

However, this single-model reuse mechanism does not directly extend to multi-model settings because the cached state produced in Eq.~\eqref{eq:prefill} is parameter-dependent and is consumed by the corresponding model during decoding in Eq.~\eqref{eq:decode}. Even for the same prompt $X$, two models $\mathcal{M}_i$ and $\mathcal{M}_j$ with $\theta_i \neq \theta_j$ produce different prefill caches, $\mathcal{C}^{(i)}_n$ and $\mathcal{C}^{(j)}_n$, making their KV states incompatible. As a result, typical multi-model systems must recompute prefill and maintain separate KV caches for each specialized model, causing redundant computation and duplicated memory usage.

\subsection{Disaggregated Serving}

Modern LLM serving systems use continuous batching~\citep{kwon2023efficient, yu2022orca, zheng2024sglang} to improve throughput by interleaving prefill and decode requests in the same scheduling loop. While effective for GPU utilization, this
couples two phases with different resource characteristics: prefill is compute-bound, whereas decoding is memory-bound and repeatedly accesses the KV cache. When a new prefill request is inserted into an active batch, it can delay ongoing decoding and cause spikes in inter-token latency (ITL), degrading interactive responsiveness.

Disaggregated serving~\citep{patel2024splitwise, zhong2024distserve} mitigates this interference by placing prefill and decode on separate instances. By isolating compute-heavy prefill from the decode loop, it stabilizes ITL and allows prefill capacity to scale independently. However, disaggregation alone does not make KV caches reusable across models. Thus, in multi-model workloads, identical prompts still require per-model prefill computation and separate KV storage, leaving cross-model redundancy unresolved.

\section{Method}
\label{sec:method}
%In this section, we present PrefillShare, including a disaggregated serving formulation with a shared prefill module and a cache-conditioned fine-tuning procedure that ensures cache compatibility between the base and specialized models.

\subsection{Definitions and Formulation}
Building on the autoregressive formulation in Section~\ref{sec:background}, we first define two operational roles for inference: a prefill module that produces a shared KV cache from the input prompt and a decode module that generates task-specific outputs by consuming the shared cache.

\paragraph{Base Prefill Module.}
Let the base prefill module be a pretrained model $\mathcal{M}_{\text{base}}$ with parameters $\theta_{\text{base}}$. The parameters $\theta_{\text{base}}$ are responsible only for the prefill phase. It processes the input $X$ once to produce a shared KV cache $\mathcal{C}_{\text{base}}$

\begin{equation}
    (\cdot, \mathcal{C}_{\text{base}}) = \mathcal{F}_{\theta_{\text{base}}}(X, \emptyset)
\end{equation}

Here $\mathcal{C}_{\text{base}}$ denotes the shared KV cache. Note that $\mathcal{M}_{\text{base}}$ computes the KV cache but does not participate in the subsequent token generation loop.

\paragraph{Specialized Decode Module.}
For each task, we construct a decode module $\mathcal{M}_{\text{dec}}$ with parameters $\theta_{\text{dec}}$. At inference time, the decode module initializes its cache with $\mathcal{C} \leftarrow \mathcal{C}_{\text{base}}$ and generates the output sequence:

\begin{equation}
    (y_t, \Delta \mathcal{C}_t) = \mathcal{F}_{\theta_{\text{dec}}}(y_{t-1}, \mathcal{C})
\end{equation}

This separation enables multiple decode modules to reuse the same shared prefill cache while preserving task-specific generation behavior.

\subsection{Cache-Conditioned Fine-Tuning}
\begin{figure}[b]
    \centering
    \includegraphics[width=0.5\linewidth]{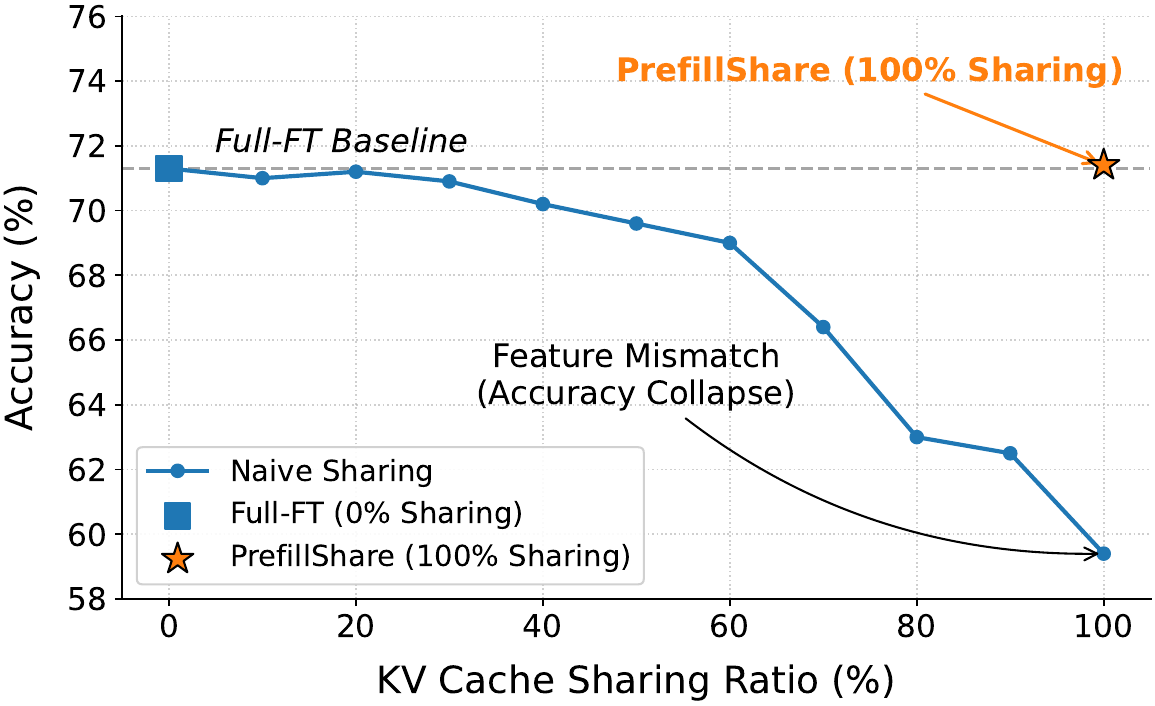}
    \caption{\textbf{Naive KV-cache sharing versus PrefillShare.} GSM8K accuracy as a function of the KV-cache sharing ratio between the base and fine-tuned models. Naive sharing collapses at high sharing ratios, whereas PrefillShare preserves near full fine-tuning (Full-FT) accuracy even at 100\% sharing.}
    \label{fig:naive_sharing}
\end{figure}

A key challenge in PrefillShare is that a task-specific decoder must generate tokens conditioned on KV caches produced by a different parameterization. In standard fine-tuning, prefill and decode are performed by the same fine-tuned model, so the decoder is trained only on its own KV-cache distribution. This setup tightly couples the decoder parameters to their own cache
representations observed during training. As shown in Fig.~\ref{fig:naive_sharing}, naively reusing an increasing fraction of KV caches from the pretrained base model introduces a distribution mismatch, causing accuracy to degrade and eventually collapse.

% In other words, naively decoding with a KV cache produced by a different model (e.g., the base model) can substantially degrade accuracy as depicted in Fig.~\ref{fig:naive_sharing}, since KV caches are model-specific and fine-tuning changes the key/value representations. 

%While prior approaches \citep{liu2024droidspeak,ye2025kvcomm} attempt to bridge distribution gaps in key/value representations across models by applying calibration based on calibration datasets or online signals, robust quality at broader scope remains difficult to guarantee, since such calibration cannot reliably cover diverse prompts and is not enforced through training under cross-model KV-cache sharing.

%PrefillShare aims to enable a task-specific decode module $\mathcal{M}_{\text{dec}}$ to generate tokens by conditioning on a shared prompt cache $\mathcal{C}_{\text{base}}$ computed by a frozen base prefill module $\mathcal{M}_{\text{base}}$. This setting requires $\mathcal{M}_{\text{dec}}$ to reliably decode from $\mathcal{C}_{\text{base}}$, even though the KV representations in $\mathcal{C}_{\text{base}}$ differ from those that would be produced under the task-specific parameters $\theta_{\text{dec}}$.

%Under PrefillShare, however, the decode module consumes $\mathcal{C}_{\text{base}}$, a shared cache computed by the frozen base prefill module $\mathcal{M}_{\text{base}}$. Directly substituting this external cache at inference time can therefore introduce feature mismatch and degrade generation quality.

To address this mismatch and enable robust shared-prefill inference, we propose cache-conditioned fine-tuning, a disaggregation-aware training procedure that explicitly trains the decode module to consume $\mathcal{C}_{\text{base}}$.
For each training example $(X, Y)$, we first compute the shared cache $\mathcal{C}_{\text{base}}$ using the base prefill module $\mathcal{M}_{\text{base}}$ from input prompt $X$, and treat the cache as a constant input so that gradients do not propagate into $\theta_{\text{base}}$. 
We then fine-tune only the task-specific decode module $\mathcal{M}_{\text{dec}}$ with parameters $\theta_{\text{dec}}$ to predict the target sequence while conditioning on the base cache:

\begin{equation}
    \mathcal{L}(\theta_{\text{dec}}) = -\sum_{t} \log P(y_t \mid y_{<t}, \mathcal{C}_{\text{base}}; \theta_{\text{dec}})
    \label{eq:cache_conditioned_ft}
\end{equation}

We train $\mathcal{M}_{\text{dec}}$ with teacher forcing, feeding the ground-truth prefix $y_{<t}$ while conditioning on the fixed cache $\mathcal{C}_{\text{base}}$. Notably, cache-conditioned fine-tuning preserves the standard next-token prediction objective, but replaces the self-generated cache with the shared base cache as the conditioning signal.

%Freezing $\mathcal{M}_{\text{base}}$ keeps the cache distribution stable, enabling multiple task-specific decode modules to be trained and served consistently on top of the same prefill representation.

By optimizing Eq.~\eqref{eq:cache_conditioned_ft}, $\mathcal{M}_{\text{dec}}$ learns to generate tokens conditioned on the KV-cache distribution produced by $\mathcal{M}_{\text{base}}$. This adaptation allows $\mathcal{M}_{\text{dec}}$ to effectively interpret the shared base cache $\mathcal{C}_{\text{base}}$, mitigating the representation mismatch that causes naive KV-cache sharing to degrade accuracy, as shown in Fig.~\ref{fig:naive_sharing}. Since the training setup matches the shared-prefill inference path, PrefillShare can reliably reuse the same base cache across heterogeneous task-specific decoders without synchronizing their parameters with the base model.

%In practical deployment, cache-conditioned fine-tuning introduces a one-time adaptation cost, whereas the serving benefits of PrefillShare recur across requests. Reducing redundant prefill computation and increasing throughput therefore amortize the additional training cost as serving continues as detailed in Appendix~\ref{appendix:adaptation_cost}.

\subsection{Disaggregated Inference Workflow}
\label{sec:workflow}

\begin{figure*}[b]
    \centering
    \includegraphics[width=\textwidth]{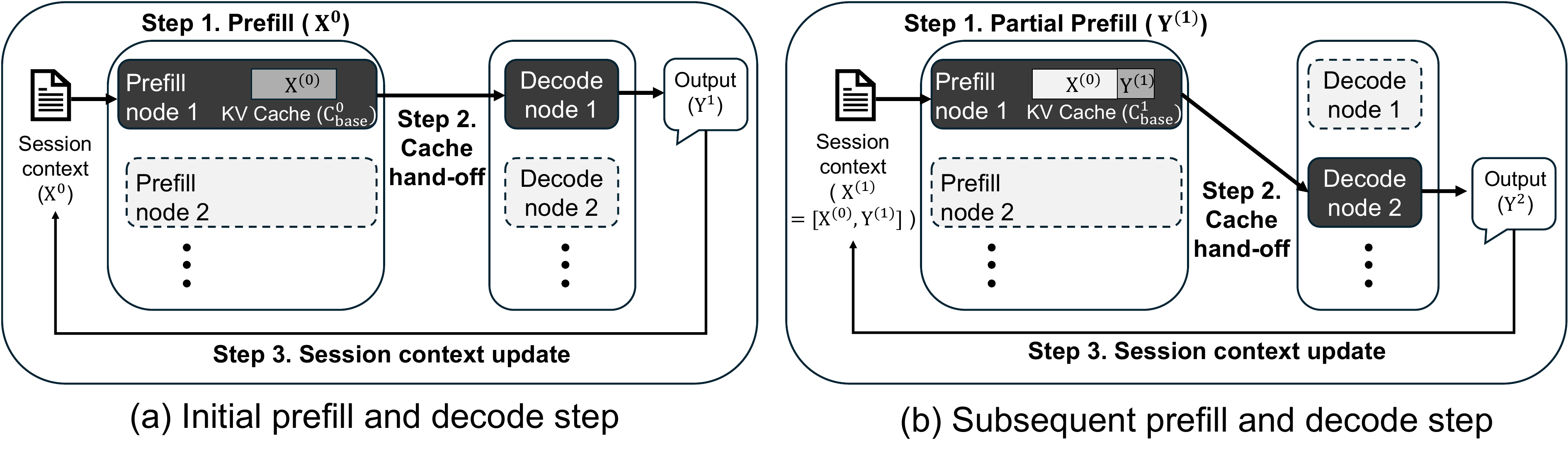}
    \caption{
        \textbf{PrefillShare disaggregated inference workflow.}
        Prefix-aware routing pins same-session requests to the same shared prefill node,
        allowing the base KV cache to be reused and incrementally extended across model
        switches. \textbf{(a)} The initial step performs full prefill and cache handoff.
        \textbf{(b)} Subsequent steps perform partial prefill and hand off the updated
        cache to another decode node.
        }
    \label{fig:inference_pipeline}
\end{figure*}

Although cache-conditioned fine-tuning enables task-specific decoders compatible
with the shared base KV cache, existing disaggregated serving systems are not
designed to preserve this cache across model switches. In prior disaggregated systems,
prefill requests are scheduled independently for each model or request, so two
invocations that share the same session prefix may be routed to different prefill
workers. This breaks prefix-cache locality: the worker handling the next model
invocation may not hold the previously computed base cache, forcing the system to
recompute the shared prefix instead of extending it incrementally.

To preserve prefix-cache locality and realize shared-prefill reuse in practice,
we implement a prefix-aware routing and cache-handoff mechanism in a
vLLM-based disaggregated serving system. The router pins invocations with the
same session prefix to the same prefill worker whenever possible, allowing the
worker to keep the base KV cache resident and extend it via partial prefill as
new tokens are appended. The resulting cache is then handed off to the selected
task-specific decoder, avoiding redundant prefill and duplicated KV-cache storage
across model switches as shown in Fig.~\ref{fig:inference_pipeline}.

Concretely, for the initial invocation, the prefill node performs full prefill
over the session context $X^{(0)}$ and constructs
$\mathcal{C}^{(0)}_{\text{base}}$, which is handed off to the first decode node
to generate $Y^{(1)}$. The generated output is appended to the session context,
forming $X^{(1)} = [X^{(0)}; Y^{(1)}]$. For subsequent model switches, the same
prefill node performs partial prefill only over the newly appended tokens,
extends the cache to $\mathcal{C}^{(1)}_{\text{base}}$, and hands the updated
cache to the next selected decode node. Repeating this process enables
heterogeneous decoders to reuse a shared prefix cache across model switches,
avoiding per-model prefix recomputation and duplicated KV-cache storage.
Appendix~\ref{appendix:memory_accounting} compares the memory required for shared-prefix KV caches and model weights under the same GPU budget.
Furthermore, even when prefix overlap is limited, any decoder can be served by any prefill worker, allowing PrefillShare to pool prefill capacity across models, balance load across prefill GPUs, and retain high serving-efficiency.

%\paragraph{Efficiency Analysis.}
%We decompose the sequence length into a shared prefix length
%$L_{\text{shared}}$ and a model-specific segment length $L_{\text{unique}}$.
%In typical disaggregated serving, each model maintains its own KV cache for the
%full sequence, yielding

%\begin{equation}
%    \text{Mem}_{\text{baseline}} =
%    O\!\left(N \cdot (L_{\text{shared}} + L_{\text{unique}})\right).
%\end{equation}

%PrefillShare stores the shared prefix cache once and maintains only
%model-specific decode segments for each decoder:

%\begin{equation}
%    \text{Mem}_{\text{PrefillShare}} =
%    O\!\left(L_{\text{shared}} + N \cdot L_{\text{unique}}\right).
%\end{equation}

%When $L_{\text{shared}}$ dominates $L_{\text{unique}}$, PrefillShare reduces the
%dominant memory cost from $O(N \cdot L_{\text{shared}})$ to
%$O(L_{\text{shared}})$, making the primary shared-cache cost independent of the
%number of decode modules. The same structure also improves responsiveness:
%typical serving must rerun prefill over the shared context when invoking a new
%model $\mathcal{M}_k$, increasing TTFT as the context grows, whereas
%PrefillShare reuses $\mathcal{C}_{\text{base}}$ and makes model switching largely
%insensitive to $L_{\text{shared}}$.

\section{Experiments}
\label{sec:experiments}

\subsection{Experimental Setup}
\label{subsec:setup}
\paragraph{Training Setup.}
We evaluate PrefillShare across multiple model families, sizes, and adaptation domains. We compare cache-conditioned fine-tuning against full fine-tuning (Full-FT) and ICaRus~\citep{woo2026icarus} using LLaMA-3.1-8B/70B ~\citep{grattafiori2024llama} and Qwen3-1.7B/8B/14B-Base~\citep{yang2025qwen3}. We fine-tune domain-specific models on MetaMathQA-40K~\citep{yu2023metamath} and DeepMath-1K~\citep{deepmath-he} for math reasoning, EvolInstruct-Code-80K~\citep{evolinstruct} for coding, and xLAM-function-calling-60K~\citep{XLAM} for tool calling. We evaluate them on GSM8K/GSM+~\citep{cobbe2021gsm8k, li-etal-2024-gsmplus}, MATH-500~\citep{hendrycks2021math}, HumanEval/HumanEval+~\citep{chen2021evaluating, liu2023evalplus},
and the Simple Python and Multiple subsets of BFCL~\citep{patil2025bfcl}, respectively. Additional training details are provided in Appendix~\ref{app:experimental details}. Appendix~\ref{appendix:adaptation_cost} further quantifies the one-time cost of cache-conditioned fine-tuning and its amortization against recurring serving-side savings.

\paragraph{Inference Setup.}
We evaluate PrefillShare in a four-agent, multi-turn serving scenario where agents are invoked sequentially over a largely shared prefix, creating frequent model switches and opportunities for prefill reuse. We use representative agentic prompting patterns, including ReAct and Reflexion, following prior LLM agent and serving studies~\citep{kim-dynamic-reasoning, woo2026icarus}. Additional inference details are provided in Appendix~\ref{app:experimental details}.

We compare against a disaggregated serving baseline with independent prefill/decode pipelines per model. The baseline uses one prefill GPU and one decode GPU per model, forming four isolated prefill/decode pairs (8 GPUs total). PrefillShare uses the same GPU budget, pooling the four prefill GPUs and sharing prefill caches across decoders. Unless stated otherwise, experiments use LLaMA-3.1-8B on a single A100 node; both systems extend vLLM's disaggregated serving pipeline. 

For each arrival rate, we sweep the maximum number of concurrent sessions, an admission-control limit that bounds the active KV-cache footprint, and report the best-performing setting.
Appendices~\ref{appendix:concurrency_ablation} and~\ref{appendix:input_length_ablation} analyze the effects of the concurrency limit and input length on serving performance, respectively.

\begin{table*}[t]
\centering
\caption{\textbf{Accuracy on math, coding, and tool-calling tasks.} LLaMA-3.1-8B and LLaMA-3.1-70B after Full-FT and PrefillShare fine-tuning, evaluated on GSM8K/GSM+, HumanEval/HumanEval+, and BFCL, respectively. PrefillShare matches full fine-tuning (Full-FT) accuracy while enabling KV-cache sharing.}
\label{tab:main_results}
\small
\setlength{\tabcolsep}{6pt}
\renewcommand{\arraystretch}{1.1}

\resizebox{\textwidth}{!}{%
\begin{tabular}{l ccc cc cc}
\toprule
\multirow{2}{*}{\textbf{Configuration}} &
\multirow{2}{*}{\textbf{KV Sharing}} &
\multicolumn{2}{c}{\textbf{Math}} &
\multicolumn{2}{c}{\textbf{Coding}} &
\multicolumn{2}{c}{\textbf{Tool calling}} \\
\cmidrule(lr){3-4}\cmidrule(lr){5-6}\cmidrule(lr){7-8}
& & \textbf{GSM8K} & \textbf{GSM+} & \textbf{HumanEval} & \textbf{HumanEval+} & \textbf{Simple Python} & \textbf{Multiple} \\
\midrule

LLaMA-3.1-8B         & Inherent      & 25.9 & 18.0 & 36.6 & 29.9 & 70.5 & 45.5 \\
\cdashline{1-8}\noalign{\vskip 0.35ex}
Full-FT             & Not Supported & 71.3 & \textbf{49.8} & 48.2 & \textbf{45.7} & 90.0 & 88.0 \\
PrefillShare (Ours) & Supported     & \textbf{71.4} & 49.3 & \textbf{48.8} & 45.1 & \textbf{90.7} & \textbf{88.5} \\
\midrule

LLaMA-3.1-70B        & Inherent      & 31.9 & 24.6 & 56.7 & 50.6 & 75.0 & 57.5 \\
\cdashline{1-8}\noalign{\vskip 0.35ex}
Full-FT             & Not Supported & 79.2 & 57.9 & \textbf{73.2} & \textbf{64.6} & \textbf{90.0} & 89.5 \\
PrefillShare (Ours) & Supported     & \textbf{80.1} & \textbf{58.9} & 72.3 & 63.8 & \textbf{90.0} & \textbf{90.5} \\
\bottomrule
\end{tabular}%
}
\end{table*}

\subsection{Training and Accuracy Evaluation}

\paragraph{Robust accuracy across tasks.}
Table~\ref{tab:main_results} demonstrates that PrefillShare achieves accuracy comparable to Full-FT across math, coding, and tool-calling tasks. Across all benchmarks, PrefillShare remains within 1\% of Full-FT and even outperforms it on GSM8K, HumanEval, and the tool-calling benchmark. We observe a similar trend for LLaMA-3.1-70B, where PrefillShare remains competitive as model scale increases. We attribute this robustness to cache-conditioned fine-tuning, which trains the decode module to interpret KV representations produced by the frozen shared prefill module. The same accuracy parity extends to additional math and coding benchmarks and to an out-of-domain evaluation, where PrefillShare remains within 0.55 points of Full-FT across all evaluated settings. Appendix~\ref{appendix:other_ood} provides the full results.
%A key distinction from naive sharing is that PrefillShare is trained to consume the base model’s prefill-side KV representations: we explicitly condition training so that the decode module can effectively leverage the KV cache produced by the frozen prefill module, rather than merely reusing caches at inference time. 

\paragraph{Scaling across model sizes.}

We next evaluate whether PrefillShare preserves the reasoning benefits of fine-tuning across model scales using Qwen3-1.7B/8B/14B-Base under a math reasoning fine-tuning setup using LoRA \citep{hu-lora}. As shown in Table~\ref{tab:qwen3_model_scale}, fine-tuning on DeepMath-1K consistently improves MATH-500 accuracy over the base models. PrefillShare exhibits the same trend, matching fine-tuning within 0.2\% across all model sizes and slightly outperforming it at 8B and 14B, so cache-conditioned fine-tuning preserves the reasoning gains of fine-tuning across model sizes.

\begin{table}[t]
\centering
\caption{\textbf{Accuracy across Qwen3 model sizes.} All models are fine-tuned on DeepMath-1K and evaluated on MATH-500. Q3-1.7B/8B/14B-Base denotes Qwen3-1.7B/8B/14B, respectively. PrefillShare achieves accuracy comparable to fine-tuning across all evaluated model sizes.}
\label{tab:qwen3_model_scale}
\small
\setlength{\tabcolsep}{6pt}
\renewcommand{\arraystretch}{1.1}

\begin{tabular}{l|ccc}
\toprule
\multicolumn{1}{c}{\multirow{2}{*}{\textbf{Configuration}}}
& \multicolumn{3}{c}{\textbf{Model}} \\
\cmidrule(lr){2-4}
\multicolumn{1}{c}{}
& \textbf{Q3-1.7B-Base}
& \textbf{Q3-8B-Base}
& \textbf{Q3-14B-Base} \\
\midrule
Baseline            & 60.6 & 69.8 & 81.6 \\
\cdashline{1-4}
Fine-tuning         & \textbf{62.8} & 72.4 & 82.8 \\
PrefillShare (Ours) & 62.6 & \textbf{72.6} & \textbf{83.0} \\
\bottomrule
\end{tabular}
\end{table}

\subsection{Serving Performance under Multi-Model Agent Workloads}
\label{sec: inference_results}

\begin{figure*}[b]
    \centering
    \includegraphics[width=1.0\textwidth]{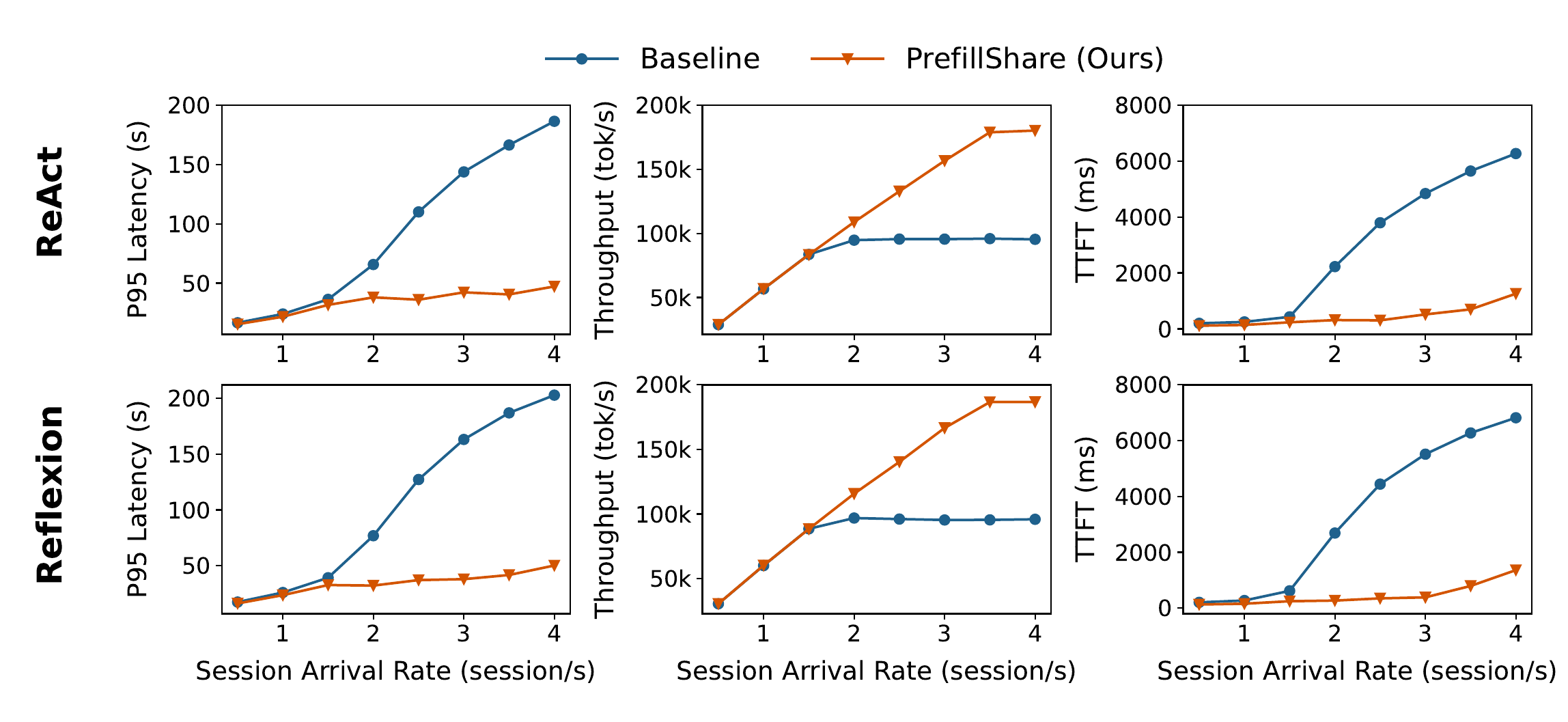}
    \caption{
        \textbf{Serving performance on multi-model agent workloads.}
        Comparison between PrefillShare and a disaggregated baseline under increasing session arrival rates for two representative agentic patterns: ReAct (top) and Reflexion (bottom). PrefillShare consistently reduces p95 latency and time to first token (TTFT) while improving throughput, with larger gains at higher arrival rates.
    }
    \label{fig:inference_results}
\end{figure*}

Fig.~\ref{fig:inference_results} compares the serving performance of PrefillShare against the disaggregated baseline under increasing session arrival rates for both ReAct and Reflexion workloads. At low load, both systems achieve similar latency and throughput. As the arrival rate increases, however, the baseline experiences rapidly growing tail latency and diminished throughput due to redundant prefill work, duplicated KV cache management across model switches, and per-model prefill workers that cannot lend capacity to one another. In contrast, PrefillShare amortizes shared-prefix processing across decoders, sustaining higher throughput while keeping tail latency low. Overall, PrefillShare achieves up to 4.1$\times$ lower p95 latency and 1.9$\times$ higher throughput on ReAct, and up to 4.5$\times$ lower p95 latency and 2.0$\times$ higher throughput on Reflexion.  Appendices~\ref{appendix:qwen}--\ref{appendix:swarm_pattern} further show that these gains persist across backbone models and agent invocation patterns. Appendix~\ref{appendix:component_ablation} separately analyzes the contributions of prefill pooling and cross-model KV reuse. Appendix~\ref{appendix:predictive_prefill_overlap} further compares PrefillShare with a predictive prefill baseline that assumes perfect next-role prediction.

\begin{table}[b]
\centering
\caption{\textbf{Serving performance of LLaMA-3.1-70B on four H100 nodes.}
One TP=8 instance per node. Baseline: a prefill and a decode node per model; PrefillShare: two shared prefill nodes.}
\label{tab:multinode}
\small
\setlength{\tabcolsep}{6pt}
\renewcommand{\arraystretch}{1.05}
\begin{tabular}{c cc cc}
\toprule
\multirow{2}{*}{\shortstack{\textbf{Session Arrival Rate}\\\textbf{(session/s)}}}
& \multicolumn{2}{c}{\textbf{Throughput (tok/s)}}
& \multicolumn{2}{c}{\textbf{TTFT (ms)}} \\
\cmidrule(lr){2-3}\cmidrule(lr){4-5}
& \textbf{Baseline}
& \textbf{PrefillShare (Ours)}
& \textbf{Baseline}
& \textbf{PrefillShare (Ours)} \\
\midrule
4  & 37.4k & \textbf{39.0k (1.04$\times$)}
   & 470   & \textbf{391 (1.20$\times$)} \\
8  & 36.6k & \textbf{56.8k (1.55$\times$)}
   & 1,559 & \textbf{717 (2.17$\times$)} \\
48 & 36.5k & \textbf{59.8k (1.64$\times$)}
   & 2,300 & \textbf{1,781 (1.29$\times$)} \\
\bottomrule
\end{tabular}
\end{table}

\paragraph{Scaling to LLaMA-3.1-70B on multiple nodes.}
\label{sec:multinode}
We further serve LLaMA-3.1-70B on four H100 nodes, each hosting one tensor-parallel instance, with two specialized models. Appendix~\ref{appendix:llama70b_multinode} provides the detailed multi-node configuration. As Table~\ref{tab:multinode} shows, the baseline saturates at about 36.5k tok/s from 4 sessions/s onward, whereas PrefillShare reaches 59.8k tok/s (1.64$\times$) and reduces TTFT by up to 2.17$\times$. The two metrics peak at different loads: the TTFT speedup is largest near the baseline's capacity limit and shrinks to 1.29$\times$ at 48 sessions/s, where decode pressure propagates back to the prefill nodes and both systems queue, whereas the throughput gain keeps growing because pooled prefill capacity keeps more work in flight. PrefillShare reduces only prefill work, so its savings surface only when decode capacity suffices.

\begin{table}[t]
\centering
\caption{\textbf{Serving performance under low prefix overlap.} An orchestrator and three specialist agents share only a 16-token (0.2\%) prefix of an 8,145-token prompt. Qwen3-14B, one 8$\times$H100 node.}
\label{tab:low_overlap}
\small
\setlength{\tabcolsep}{4pt}
\renewcommand{\arraystretch}{1.05}
\begin{tabular}{ll cccc}
\toprule
\multicolumn{2}{l}{\textbf{Session Arrival Rate (session/s)}} 
& \textbf{0.010} & \textbf{0.018} & \textbf{0.026} & \textbf{0.032} \\
\midrule
\multirow{2}{*}{p95 latency (s)}
 & Baseline            & \textbf{5.54} & 9.33 & 31.42 & 36.46 \\
 & PrefillShare (Ours) & 5.59 (0.99$\times$) & \textbf{8.13 (1.15$\times$)} & \textbf{12.07 (2.60$\times$)} & \textbf{15.46 (2.36$\times$)} \\
\cdashline{1-6}\noalign{\vskip 0.35ex}
\multirow{2}{*}{Throughput (tok/s)}
 & Baseline            & \textbf{17.5k} & 26.5k & 18.9k & 17.0k \\
 & PrefillShare (Ours) & 17.4k (0.99$\times$) & \textbf{27.6k (1.04$\times$)} & \textbf{33.9k (1.79$\times$)} & \textbf{36.7k (2.16$\times$)} \\
\bottomrule
\end{tabular}
\end{table}

\paragraph{Serving under low prefix overlap.}
\label{sec:low_overlap}
We further evaluate PrefillShare under a supervisor pattern commonly used in real-world agent workflows~\citep{hong2024metagpt,wu2024autogen}, where a main agent delegates tasks to subagents and consequently shares less prefix across model switches. As detailed in Appendix~\ref{appendix:orchestrator_pattern}, only 0.2\% of the prefix is reusable across roles in this workload. As shown in Table~\ref{tab:low_overlap}, even under this near-zero-overlap setting, PrefillShare reduces p95 latency by 62\% and improves throughput by 79\% at 0.026 sessions/s. The gain primarily comes from better utilization of prefill capacity. In our measurements in Appendix~\ref{appendix:orchestrator_analysis}, the max--min worker-load ratio decreases from 42.2$\times$ under role-affine placement to 2.07$\times$ with shared prefill workers. Concentrating requests on a dedicated worker also induces repeated prefill computation, with the orchestrator worker performing 7.55$\times$ its session-local no-eviction prefill work; pooling reduces the aggregate recomputation amplification to 1.06$\times$. 
% We further separates the two effects in Appendix~\ref{appendix:component_ablation}, showing that pooling alone improves throughput by 84\%, while exact cross-model KV reuse provides an additional 20\% improvement.

\subsection{Comparisons to Prior Work}
\label{sec:comparisons_prior_work}

\begin{figure}[b]
    \centering
    \begin{minipage}{0.47\columnwidth}
        \centering
        \footnotesize
        \setlength{\tabcolsep}{3pt}
        \renewcommand{\arraystretch}{1.08}
        \begin{tabular}{lccc}
        \toprule
        \textbf{Configuration} & \textbf{Math} & \textbf{Coding} & \textbf{Tool Calling} \\
        \midrule
        Full-FT             & 60.6 & \textbf{47.0} & 89.0 \\
        \cdashline{1-4}
        ICaRus              & 59.1 & 46.1 & 88.3 \\
        PrefillShare (Ours) & \textbf{60.4} & \textbf{47.0} & \textbf{89.6} \\
        \bottomrule
        \end{tabular}
    \end{minipage}\hfill
    \begin{minipage}{0.47\columnwidth}
        \centering
        \includegraphics[width=\linewidth]{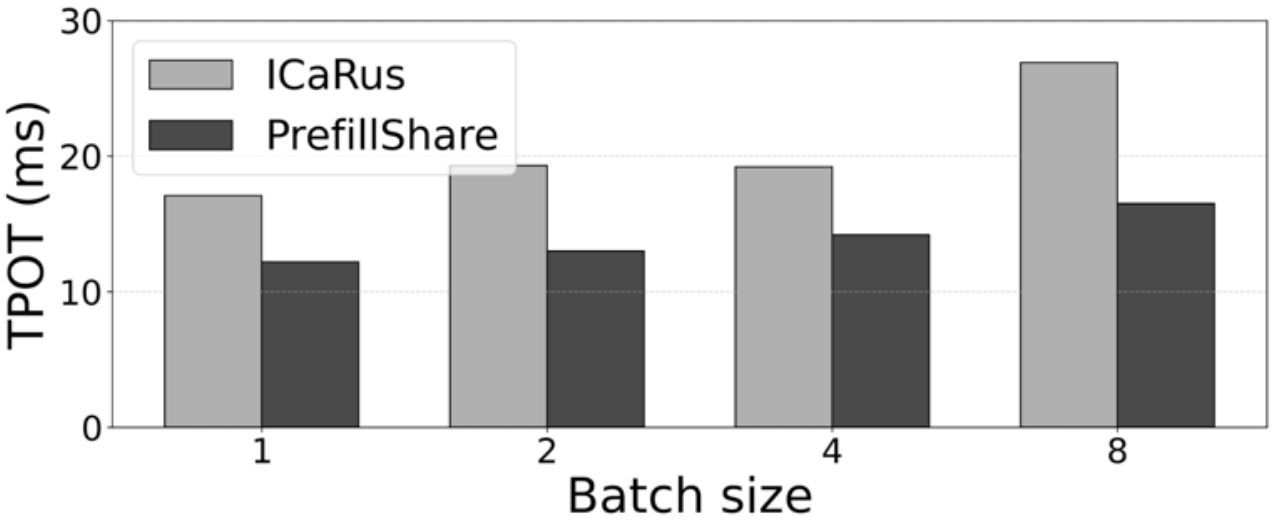}
    \end{minipage}
    \caption{
    \textbf{Comparison between PrefillShare and ICaRus on LLaMA-3.1-8B.}
    \textbf{(Left)} Accuracy across math (GSM8K/GSM+), coding
    (HumanEval/HumanEval+), and tool calling (BFCL Simple Python/Multiple)
    tasks.
    \textbf{(Right)} Time per output token (TPOT) measured on a single A100 GPU
    with input length 1K.
    PrefillShare outperforms ICaRus in both task accuracy and serving efficiency.
}
    \label{fig:prior_work_llama}
\end{figure}

We compare PrefillShare with ICaRus, a prior approach to robust KV-cache reuse across fine-tuned models. As shown in
Fig.~\ref{fig:prior_work_llama}, PrefillShare achieves higher accuracy than
ICaRus across all evaluated task categories while also improving serving
efficiency. In particular, PrefillShare improves math accuracy by 1.3\% over ICaRus, matches Full-FT on coding, and achieves the highest accuracy on tool calling. On the systems side, PrefillShare reduces TPOT from 17.1 ms to 12.2 ms at batch size 1, corresponding to a 28.7\% reduction. The benefit becomes larger at higher load: at batch size 8, TPOT decreases from 26.9 ms to 16.5 ms, yielding a 38.7\% reduction.

The difference comes from where KV-cache sharing is enforced. ICaRus keeps the frozen base model on the decoding path so that KV caches remain compatible across fine-tuned variants. This enables reuse of KV states generated during decoding, but the base model and its task-specific adaptation must run together at every decode step; adapter-based tuning such as LoRA reduces this cost, yet limits adaptation capacity relative to full fine-tuning and adds adapter overhead to decoding. PrefillShare instead uses the base model only for shared prefill: the base model produces the shared KV cache once, and fully fine-tuned task-specific decoders consume this cache for generation. This
separation preserves full fine-tuning capacity while removing adapter overhead from decoding, which explains PrefillShare's gains in both accuracy and TPOT. Appendix~\ref{appendix:icarus_matched} compares PrefillShare and ICaRus under the same LoRA configuration to control for adaptation capacity. Appendix~\ref{appendix:kvcomm_droidspeak} provides a qualitative comparison with DroidSpeak and KVCOMM.

\section{Related Work}

\paragraph{Efficient LLM Serving and KV Management.}
vLLM's PagedAttention~\citep{kwon2023efficient} introduces block-based KV-cache management that reduces fragmentation and improves batching efficiency. Prefix caching~\citep{gim2024prompt} reuses KV states for repeated prompt prefixes, and SGLang's RadixAttention~\citep{zheng2024sglang} organizes them in a radix tree for cross-request reuse.

\paragraph{Disaggregated Serving.}
Disaggregated serving addresses the prefill--decode interference of continuous batching~\citep{yu2022orca} by separating the two phases. DistServe~\citep{zhong2024distserve} assigns each phase to dedicated resources to improve latency stability, while Splitwise~\citep{patel2024splitwise} formalizes phase separation as a system-level optimization for goodput. Mooncake~\citep{qin2026mooncake} further decouples KV-cache management from execution in large-scale serving.

\paragraph{Multi-Model and Agentic Inference.}
Agentic systems combine reasoning, planning, and tool use~\citep{schick2023toolformer, shinn2023reflexion, wang2023plan, wei2022cot, yao2022react}. Multi-agent frameworks such as
LangChain~\citep{langchain2023} and AutoGen~\citep{wu2024autogen} assign specialized roles to agents, while heterogeneous orchestration systems such as ToolOrchestra~\citep{su2025toolorchestra} combine
models and tools. DroidSpeak and KVCOMM address related inefficiencies via layer-wise and anchor-based KV reuse, respectively~\citep{liu2024droidspeak, ye2025kvcomm}, whereas PrefillShare targets shared-prefill reuse across task-specific decoders.
\section{Conclusion}

We proposed PrefillShare, a disaggregated serving algorithm that eliminates redundant prefill computation and duplicated KV caches in multi-model LLM workloads by separating inference into a shared prefill module and task-specific decode modules. Enabled by cache-conditioned fine-tuning, which aligns specialized decoders with a frozen shared prefill module, our approach preserves task accuracy while substantially improving tail latency and throughput across diverse tasks, model scales, and agent workloads, even when agents share almost no prefix. Shared-prefill execution is thus a promising foundation for scalable multi-model and agentic LLM serving.

\bibliography{example_paper}
\bibliographystyle{iclr2027_conference}

%%%%%%%%%%%%%%%%%%%%%%%%%%%%%%%%%%%%%%%%%%%%%%%%%%%%%%%%%%%%%%%%%%%%%%%%%%%%%%%
%%%%%%%%%%%%%%%%%%%%%%%%%%%%%%%%%%%%%%%%%%%%%%%%%%%%%%%%%%%%%%%%%%%%%%%%%%%%%%%
% APPENDIX
%%%%%%%%%%%%%%%%%%%%%%%%%%%%%%%%%%%%%%%%%%%%%%%%%%%%%%%%%%%%%%%%%%%%%%%%%%%%%%%
%%%%%%%%%%%%%%%%%%%%%%%%%%%%%%%%%%%%%%%%%%%%%%%%%%%%%%%%%%%%%%%%%%%%%%%%%%%%%%%
\clearpage
\appendix
\raggedbottom
% \onecolumn
\section*{Appendices}

\section{Experimental Details}\label{app:experimental details}

\subsection{Training Setup}
\label{app:training setup}

All training experiments reported in Table~\ref{tab:main_results} were carried out on a single node with 8 A100-SXM4-80G GPUs. For fine-tuning, we used the LMFlow \citep{diao2024lmflowextensibletoolkitfinetuning} framework with three datasets: MetaMathQA-40K \citep{yu2023metamath}, EvolInstruct-Code-80K \citep{evolinstruct}, and xLAM-function-calling-60K \citep{XLAM}.

Fine-tuning tasks across all datasets were consistently performed with a batch size of 1, gradient accumulation over 16 steps and a maximum sequence length of 1024 for one epoch. Learning rates were selected via a grid search over $\{1 \times 10^{-4}, 1 \times 10^{-5}, 2 \times 10^{-5}, 5 \times 10^{-5}, 2 \times 10^{-6}, 5 \times 10^{-6} \}$ for each dataset. AdamW with $\beta_1=0.9$ and $\beta_2=0.999$ was used as the optimizer for fine-tuning, together with a weight decay of 0.1 and a warmup ratio of 0.03. There were no explicit regularization setups, including gradient clipping or dropout.

\subsection{Inference Setup}
\label{appendix:detailed_inference_setup}

\paragraph{Workloads and measurement.}
All inference experiments follow the serving configuration described in
Section~\ref{subsec:setup}. For a fair comparison, both systems are evaluated
under the same workload and generation settings. We replay the same session-level
traces for the disaggregated baseline and PrefillShare, so each system receives
the same session arrivals and agent invocation sequence.

For the ReAct~\citep{yao2022react} and Reflexion~\citep{shinn2023reflexion}
agents, each session contains four agents invoked in a round-robin order for
four rounds, resulting in 16 model invocations per session. The input and output
token lengths of each invocation are fixed according to the token-length
statistics reported in prior agent-serving work~\citep{kim-dynamic-reasoning,
woo2026icarus}. Each session proceeds synchronously: after receiving a model
response, it immediately issues the next model invocation until all turns are
completed, without injecting additional delays between model calls. Thus, the
session arrival rate specifies how frequently new sessions enter the system,
rather than the rate of individual model invocations.

Across agent transitions, both systems append generated responses to the session
context using the same prompt format. Latency and throughput are measured at the
client-facing proxy to capture the end-to-end serving behavior observed by
requests.

\paragraph{Multi-node configuration.}
\label{appendix:llama70b_multinode}
The four H100 nodes are connected by InfiniBand, and each hosts one TP=8 instance so that decode work is spread over all eight GPUs of the node. The baseline dedicates one prefill and one decode node to each of the two specialized models, whereas PrefillShare uses two shared prefill nodes and two task-specific decode nodes. Batching, chunked prefill, prefix caching, decode capacity, and session arrivals are identical for both systems, and the ReAct workflow follows the single-node setup above. Input contexts are longer than in a single node workload, so absolute throughput is not comparable across the two settings.

\paragraph{Low-overlap orchestrator workload.}
\label{appendix:orchestrator_pattern}
We construct a supervisor-style workload in which a main agent delegates role-specific tasks to three subagents in the sequence
\texttt{main $\rightarrow$ A $\rightarrow$ main $\rightarrow$ B $\rightarrow$ main $\rightarrow$ C $\rightarrow$ main}.
Each role maintains an independent history, and the main agent provides each subagent only with its role-specific instruction, substantially reducing the prefix shared across model switches. Per-step input and output lengths follow TraceLab~\citep{zhu2026tracelab}, a trace of approximately 4,300 real Claude Code and Codex sessions.

The resulting prompts share only 30 tokens across roles out of 8,145 tokens (0.4\%). With the 16-token KV-cache block granularity used by our serving system, only one block remains reusable, corresponding to 0.20\% of the prompt. Both systems serve Qwen3-14B-Base on one 8$\times$H100 node with four prefill and four decode workers. We use identical batching, chunked prefill, prefix caching, decode capacity, and session arrivals for the baseline and PrefillShare.

\paragraph{KV transfer at high concurrency.}
\label{appendix: vLLM implementation}
Our prototype is implemented on top of vLLM's disaggregated serving pipeline.
In the common case, KV caches produced by prefill workers are transferred
to decode workers and kept resident in GPU memory, where they are directly consumed during autoregressive decoding.

At high concurrency, when many sessions are concurrently active, the aggregate GPU-resident KV footprint on the decode side can temporarily exceed the available memory capacity.
In such cases, vLLM stages a subset of KV caches in CPU memory and reloads them to the GPU when the corresponding sessions are scheduled for decoding.
This swap-like staging and reload behavior increases CPU--GPU data movement, which can increase latency and reduce throughput at high concurrency.

We emphasize that this behavior is an implementation-level mechanism in vLLM for handling transient KV memory pressure, rather than an inherent limitation of shared-prefill execution.
In principle, alternative designs could mitigate overflow-induced staging via stricter admission control, decode-to-prefill backpressure, or per-session reservation of GPU-resident KV buffers.
Exploring such alternatives is orthogonal to PrefillShare and left for future system optimizations.

\section{Robustness Across Backbones and Agent Invocation Patterns}
\label{appendix:robustness}

\begin{figure}[t]
    \centering
    \includegraphics[width=1.0\textwidth]{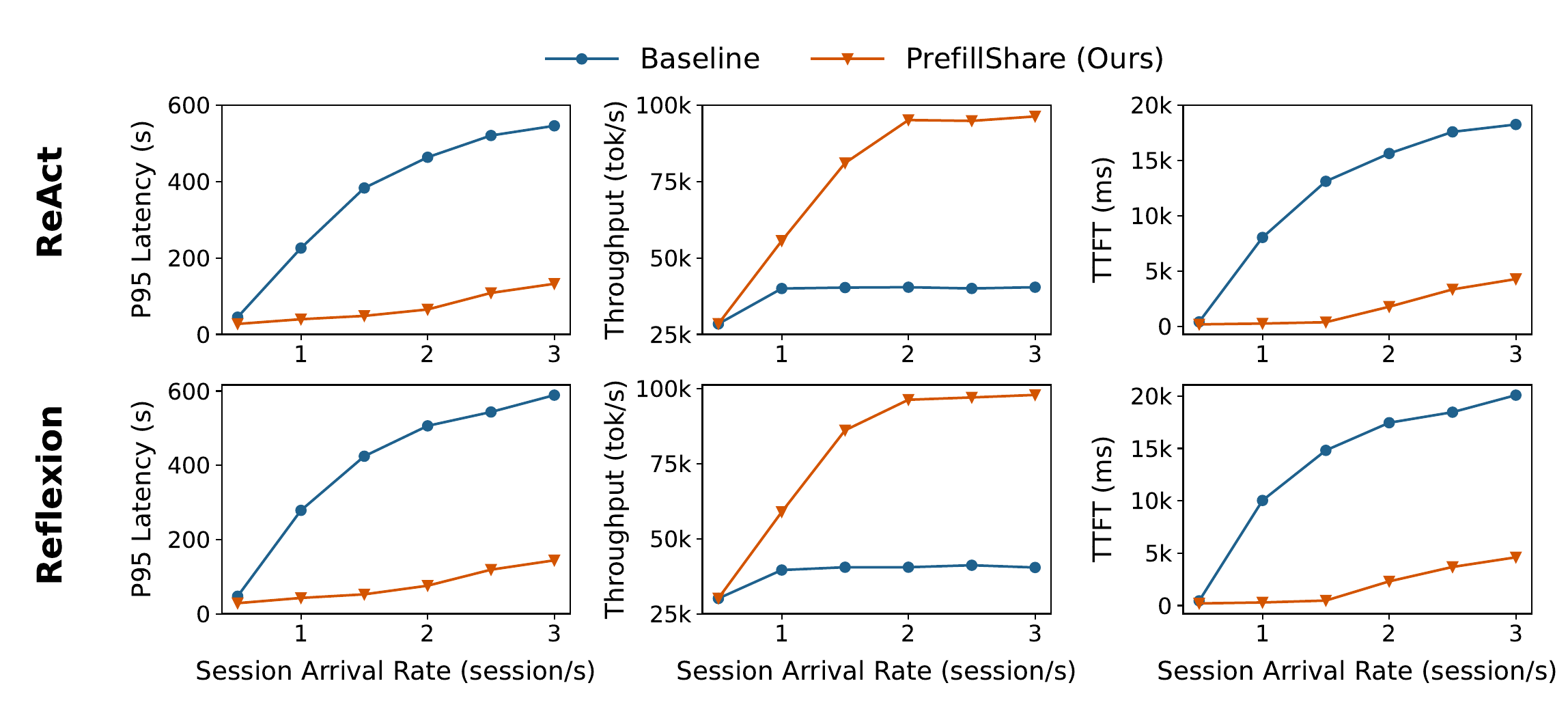}
    \caption{\textbf{Serving performance with Qwen3-14B.}
        We replicate the setup of Fig.~\ref{fig:inference_results}, replacing the LLaMA-3.1-8B backbone with Qwen3-14B while keeping all other settings identical.
        Results are shown for ReAct (top) and Reflexion (bottom), reporting p95 end-to-end latency, throughput, and time-to-first-token (TTFT).
    }
    \label{fig:appendix_inference_results}
\end{figure}

\begin{figure}[b]
  \centering
  \includegraphics[width=\columnwidth]{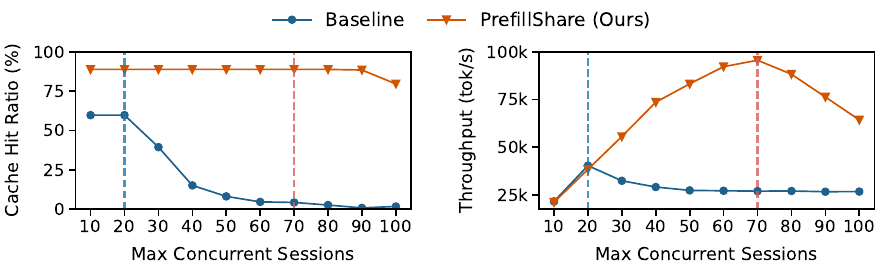}
  \caption{\textbf{Concurrency ablation with Qwen3-14B.}
      This figure mirrors the setup of Fig.~\ref{fig:inflight_hit_throughput}, with the backbone model replaced by Qwen3-14B.
      Left: prefix cache hit ratio (\%). Right: throughput (tok/s).
  }
  \label{fig:appendix_hit_throughput}
\end{figure}

\subsection{Qwen3-14B Backbone}
\label{appendix:qwen}

To examine whether the serving behavior observed in Section~\ref{sec: inference_results} depends on a specific backbone model, we repeat the same experiments using Qwen3-14B in place of LLaMA-3.1-8B.
All workload configurations, scheduling policies, and evaluation protocols remain identical to those used in the main experiments. As shown in Fig.~\ref{fig:appendix_inference_results} and Fig.~\ref{fig:appendix_hit_throughput}, PrefillShare continues to achieve substantially higher throughput and lower tail latency than the baseline, while maintaining a high prefix cache hit ratio over a wide concurrency range.

\subsection{Randomized Swarm Pattern}
\label{appendix:swarm_pattern}

\begin{table}[t]
\centering
\caption{\textbf{Serving performance under a randomized four-agent swarm pattern.}
TTFT and throughput comparison between the disaggregated baseline and
PrefillShare on LLaMA-3.1-8B with four sequential agents on a single A100 node.
Each agent randomly hands off the request to another agent after completing its
task segment. PrefillShare consistently reduces time-to-first-token (TTFT) and improves throughput
under higher load.}
\label{tab:swarm_pattern}
\small
\setlength{\tabcolsep}{5pt}
\renewcommand{\arraystretch}{1.12}

\begin{tabular}{c cc cc}
\toprule
\multirow{2}{*}{\shortstack{\textbf{Session Arrival Rate}\\\textbf{(session/s)}}}
& \multicolumn{2}{c}{\textbf{TTFT (ms)}}
& \multicolumn{2}{c}{\textbf{Throughput (tok/s)}} \\
\cmidrule(lr){2-3}\cmidrule(lr){4-5}
& \textbf{Baseline}
& \textbf{PrefillShare (Ours)}
& \textbf{Baseline}
& \textbf{PrefillShare (Ours)} \\
\midrule
0.5 & 152 & \textbf{76 (2.0$\times$)} & 7,314  & \textbf{7,319 (1.0$\times$)} \\
1.5 & 280 & \textbf{81 (3.5$\times$)} & 21,094 & \textbf{21,280 (1.0$\times$)} \\
2.5 & 733 & \textbf{94 (7.8$\times$)} & 28,206 & \textbf{34,272 (1.2$\times$)} \\
3.5 & 741 & \textbf{141 (5.3$\times$)} & 28,325 & \textbf{45,919 (1.6$\times$)} \\
\bottomrule
\end{tabular}
\end{table}

We further evaluate PrefillShare under a randomized four-agent swarm pattern on
LLaMA-3.1-8B with a single A100 node, using the same setup as
Appendix~\ref{appendix:detailed_inference_setup} except for the agent invocation
pattern. In other words, unlike the fixed round-robin patterns in Section~\ref{sec: inference_results},
the swarm workload lets each agent randomly hand off the request to another
agent after completing its task segment, while preserving a shared and growing
session context across agents.

Table~\ref{tab:swarm_pattern} reports TTFT and throughput, showing that
PrefillShare consistently reduces TTFT across all session arrival rates,
achieving up to 7.8$\times$ lower TTFT. These results show that the benefit of PrefillShare is not specific to a fixed agent invocation pattern. Regardless of whether model transitions follow a fixed sequence or a randomized swarm, the baseline must repeatedly recompute prefill and maintain duplicated KV caches across specialized models. PrefillShare removes this structural redundancy by sharing the base prefix cache across model handoffs, making its efficiency gains largely independent of the particular agent invocation order.

\section{Analysis of Serving Mechanisms and Resource Utilization}
\label{appendix:serving_analysis}

\subsection{Attribution of Prefill Pooling and Cross-Model KV Reuse}
\label{appendix:component_ablation}

PrefillShare combines two mechanisms that can independently improve serving efficiency: sharing prefill capacity across models and reusing KV caches across model switches. To separate their contributions, we evaluate an intermediate configuration that enables the shared prefill pool, prefix caching, and prefix-aware routing while disabling cross-model KV reuse. In this configuration, any model can use any prefill worker, but each model still recomputes its own prefix.

\begin{table}[t]
\centering
\caption{\textbf{Attribution of prefill pooling and cross-model KV reuse.}
All configurations use the same Qwen3-14B 4P+4D deployment with 64 sessions at 0.5 sessions/s.}
\label{tab:component_ablation}
\small
\setlength{\tabcolsep}{3.5pt}
\renewcommand{\arraystretch}{1.08}
\begin{tabular}{lrrrrr}
\toprule
\textbf{Configuration}
& \textbf{Mean TTFT}
& \textbf{p95}
& \textbf{Throughput}
& \textbf{Local prefill}
& \textbf{APC hit} \\
& \textbf{(s)}
& \textbf{(s)}
& \textbf{(tok/s)}
& \textbf{(M tok)}
& \textbf{(\%)} \\
\midrule
Baseline (per-model prefill)
& 8.28 & 322.60 & 28.68k & 9.56 & 20.0 \\
Shared pool, no cross-model reuse
& 3.05 & 132.09 & 52.71k & 4.92 & 58.8 \\
PrefillShare
& 1.63 & 87.02 & 63.13k & 0.68 & 94.4 \\
\bottomrule
\end{tabular}
\end{table}

Pooling alone substantially improves serving efficiency: compared with the per-model baseline, the shared-pool configuration improves throughput by 84\% and reduces p95 latency by 59\%. Exact cross-model KV reuse provides an additional gain beyond pooling. Moving from the shared-pool configuration to PrefillShare further improves throughput by 20\% and reduces p95 latency by 34\%, while increasing the APC hit rate from 58.8\% to 94.4\% and reducing local prefill from 4.92M to 0.68M tokens.

Cross-model cache reuse cannot be obtained through scheduling alone because KV states produced by independently fine-tuned models are not interchangeable. Cache-conditioned fine-tuning instead trains each task-specific decoder to consume the shared base-model KV cache, enabling exact reuse without recomputing the shared prefix. Fig.~\ref{fig:naive_sharing} shows the corresponding effect on accuracy: naive KV sharing fails, whereas PrefillShare preserves Full-FT accuracy. These results show that prefill pooling and exact cross-model KV reuse make complementary contributions to PrefillShare.

\subsection{Predictive Prefill Overlap}
\label{appendix:predictive_prefill_overlap}

We evaluate an optimistic predictive-overlap baseline that assumes the next role is known before the current one finishes decoding. While agent $i$ decodes, the baseline prefills the available prefix for the next agent on the worker that will serve it, capped at two warmups per rank. Agent $i$'s own output and the following tool result cannot be prefetched because neither exists yet. All three configurations use the same Qwen3-14B 4P+4D deployment, 64 sessions at 0.5 sessions/s.

\begin{table}[b]
\centering
\caption{\textbf{Predictive prefill overlap.}
Comparison with a predictive-overlap baseline that assumes perfect next-role prediction.}
\label{tab:predictive_prefill_overlap}
\small
\setlength{\tabcolsep}{3.5pt}
\renewcommand{\arraystretch}{1.08}
%\resizebox{\linewidth}{!}{%
\begin{tabular}{lrrrrr}
\toprule
\textbf{Configuration}
& \textbf{Mean TTFT}
& \textbf{p95}
& \textbf{Throughput}
& \textbf{Local prefill}
& \textbf{APC hit} \\
& \textbf{(s)}
& \textbf{(s)}
& \textbf{(tok/s)}
& \textbf{(M tok)}
& \textbf{(\%)} \\
\midrule
Baseline (per-model prefill)
& 8.28 & 322.60 & 28.68k & 9.56 & 20.0 \\
Predictive prefill overlap
& 7.81 & 331.88 & 28.09k & 9.72 & 40.7 \\
PrefillShare
& \textbf{1.63} & \textbf{87.02} & \textbf{63.13k} & \textbf{0.68} & \textbf{94.4} \\
\bottomrule
\end{tabular}%
%}
\end{table}

Predictive prefill overlap reduces mean TTFT by 6\% and increases the prefix-cache hit rate from 20.0\% to 40.7\%, but does not improve throughput or tail latency at this load. The overlap moves prefill work earlier without removing it, and local prefill increases from 9.56M to 9.72M tokens. The overlap is useful when prefill capacity is not the bottleneck: at 0.05 sessions/s, it reduces TTFT by 31\%. This advantage disappears as the prefill workers saturate because the prefetched work still consumes the same serving capacity. PrefillShare removes the redundant work instead, reducing local prefill to 0.68M tokens and continuing to improve throughput under load.

\subsection{Prefill-Worker Load under Low Prefix Overlap}
\label{appendix:orchestrator_analysis}

We analyze why pooling provides a large serving benefit in the low-overlap workload. Each workflow invokes the main agent four times but each specialist only once. Under the baseline, each role is bound to one prefill worker, so the asymmetric invocation frequency concentrates requests on the main-agent worker while leaving capacity on the specialist workers underutilized. PrefillShare removes this role affinity and allows requests from any role to use any of the four prefill workers.

We measure the resulting worker-level prefill load at 0.02625 sessions/s, corresponding to the 0.026 sessions/s setting in Table~\ref{tab:low_overlap}. Prefill work is measured as prefix-cache query tokens minus prefix-cache hit tokens, i.e., the number of tokens actually computed by each prefill worker.

\begin{table}[t]
\centering
\caption{\textbf{Prefill-worker load under low prefix overlap.}
Prefill tokens computed by each worker at 0.02625 sessions/s. Baseline workers are bound to individual roles, whereas PrefillShare workers are shared across roles.}
\label{tab:low_overlap_load}
\small
\setlength{\tabcolsep}{4pt}
\renewcommand{\arraystretch}{1.08}
\begin{tabular}{llrr}
\toprule
\textbf{Worker}
& \textbf{Baseline assignment}
& \textbf{Baseline}
& \textbf{PrefillShare} \\
& & \textbf{prefill tokens} & \textbf{prefill tokens} \\
\midrule
Worker 1 & Main agent   & 4,668,967 & 391,369 \\
Worker 2 & Specialist A &   254,739 & 550,692 \\
Worker 3 & Specialist B &   168,097 & 265,681 \\
Worker 4 & Specialist C &   110,625 & 390,771 \\
\midrule
Total    &               & 5,202,428 & 1,598,513 \\
\bottomrule
\end{tabular}
\end{table}

The baseline exhibits severe load imbalance: the busiest worker processes 42.2$\times$ as many prefill tokens as the least-loaded worker. PrefillShare reduces this max--min ratio to 2.07$\times$ by distributing role requests across the shared worker pool. The concentrated baseline load also causes repeated prefill computation. The main-agent worker performs 7.55$\times$ the prefill work required if each session-local prefix were computed once without eviction, whereas PrefillShare reduces the aggregate recomputation amplification to 1.06$\times$.

This behavior explains why pooling remains effective even when little prefix can be reused across models. In the baseline, requests queue behind the overloaded role-affine worker while prefill capacity on other workers remains available. PrefillShare makes this otherwise idle capacity available to the queued requests, reducing both queueing and repeated prefill computation. Static GPU reallocation does not provide the same flexibility when all roles must remain resident: removing a role from a GPU incurs model-loading latency when the role is next invoked~\citep{fu2024serverlessllm}, while the most heavily loaded role can change across tasks and over time~\citep{sun2024llumnix,duan2024muxserve,li2023alpaserve}.

\subsection{Effect of the Concurrency Limit}
\label{appendix:concurrency_ablation}

To isolate the effect of cache pressure, we introduce and ablate the max concurrent session limit, an admission-control knob that caps the number of sessions kept active in the system at a time; sessions beyond the limit remain queued until capacity becomes available. This limit controls the active KV-cache footprint and prevents unbounded session admission from causing excessive cache pressure. As concurrency increases, the disaggregated baseline must retain separate prefix KV caches for each specialized model, increasing memory pressure, cache evictions, and prefill recomputation.

\begin{figure}[b]
  \centering
  \includegraphics[width=\columnwidth]{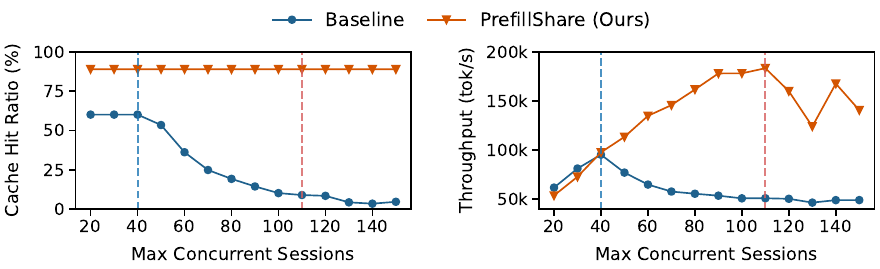}
  \caption{\textbf{Effect of the concurrency limit.} Prefix cache hit ratio (left) and throughput (right) under varying maximum concurrent sessions.
The baseline degrades beyond $\approx$40 sessions as reduced prefix reuse lowers both metrics, whereas PrefillShare sustains higher throughput over a wider range, with high-concurrency saturation driven by handoff overheads.}
  \label{fig:inflight_hit_throughput}
\end{figure}

For Fig.~\ref{fig:inference_results}, we sweep this concurrency limit and report the best-performing configuration for each setting. In Fig.~\ref{fig:inflight_hit_throughput}, we fix the ReAct arrival rate to 4 sessions/s and vary the max concurrent sessions. The baseline begins to degrade beyond roughly 40 concurrent sessions: its prefix-cache hit ratio peaks at about 60\% and then drops sharply, which reduces throughput. In contrast, PrefillShare maintains a hit ratio near 89\% across concurrency levels by storing the shared prefix cache only once, allowing throughput to scale over a wider range.

At extreme concurrency, around 110 concurrent sessions, PrefillShare throughput eventually saturates and declines even though the hit ratio remains stable. This indicates that the bottleneck shifts from prefix-cache reuse to decode-side KV pressure and cache-handoff overhead in our vLLM prototype as detailed in Appendix~\ref{app:experimental details}. The same qualitative trend holds for Qwen3-14B under identical settings, with additional results in Fig.~\ref{fig:appendix_hit_throughput}.

\subsection{Effect of Input Length}
\label{appendix:input_length_ablation}

We further study how input length affects the serving benefit of PrefillShare. We evaluate LLaMA-3.1-8B with four sequential agents on a single A100 node under two input lengths: a short context of 64 tokens and a long context of 4096 tokens. As shown in Table~\ref{tab:input_length_ablation}, PrefillShare consistently reduces TTFT across both input lengths. For short inputs, PrefillShare reduces TTFT from 0.52s to 0.27s, corresponding to a 1.9$\times$ reduction, while throughput remains unchanged at 1.0$\times$ because the prefill cost is relatively small. For long inputs, the benefit becomes much more pronounced: PrefillShare reduces TTFT from 20.18s to 6.11s, achieving a 3.3$\times$ reduction, and improves throughput from 40,218 tok/s to 88,323 tok/s, a 2.2$\times$ increase. These results confirm that PrefillShare is particularly effective for long-context workloads, where shared-prefix reuse eliminates substantial redundant prefill computation and KV cache memory. Together with the low-overlap analysis, this ablation separates the two regimes: the benefit from exact cache reuse grows with the reusable prefix, while prefill pooling remains useful even when the reusable prefix is small.

\begin{table}[t]
\centering
\caption{\textbf{Effect of input length on serving performance.}
Ablation study on time-to-first-token (TTFT) and throughput for LLaMA-3.1-8B
with four sequential agents on a single A100 node. PrefillShare provides larger
benefits for longer inputs, where shared-prefix reuse avoids more redundant
prefill computation and KV-cache storage.}
\label{tab:input_length_ablation}
\small
\setlength{\tabcolsep}{5pt}
\renewcommand{\arraystretch}{1.12}

\begin{tabular}{c cc cc}
\toprule
\multirow{2}{*}{\textbf{Input Length}}
& \multicolumn{2}{c}{\textbf{TTFT (s)}}
& \multicolumn{2}{c}{\textbf{Throughput (tok/s)}} \\
\cmidrule(lr){2-3}\cmidrule(lr){4-5}
& \textbf{Baseline}
& \textbf{PrefillShare (Ours)}
& \textbf{Baseline}
& \textbf{PrefillShare (Ours)} \\
\midrule
64   & 0.52  & \textbf{0.27 (1.9$\times$)} & 2,515  & \textbf{2,515 (1.0$\times$)} \\
4096 & 20.18 & \textbf{6.11 (3.3$\times$)} & 40,218 & \textbf{88,323 (2.2$\times$)} \\
\bottomrule
\end{tabular}
\end{table}

\section{Training Cost, Memory Accounting, and Other Benchmarks}
\label{appendix:deployment_analysis}

\subsection{Cache-Conditioned Fine-Tuning Cost and Amortization}
\label{appendix:adaptation_cost}

A method that shifts cost from serving to training should account for both sides. We therefore measure the one-time cost of cache-conditioned fine-tuning and compare it with the recurring serving-side savings. One epoch requires 2.0, 3.0, and 7.1 H100 GPU-hours for math, tool calling, and coding, respectively, for a total of 12.1 GPU-hours across the three task-specific decoders.

\begin{table}[b]
\centering
\caption{\textbf{Training cost and recurring serving-side savings.}
The training cost is measured once for the three task-specific decoders. Serving-side measurements use the Qwen3-14B 4P+4D workload from Appendix~\ref{appendix:component_ablation}.}
\label{tab:adaptation_cost}
\small
\setlength{\tabcolsep}{5pt}
\begin{tabular}{lll}
\toprule
\textbf{Quantity} & \textbf{When paid} & \textbf{Measured} \\
\midrule
Cache-conditioned fine-tuning
& Once
& 2.0 / 3.0 / 7.1 H100 GPU-h \\
Redundant prefill compute
& Every workflow
& 9.56 $\rightarrow$ 0.68 M tokens \\
End-to-end throughput
& Every workflow
& 28.68k $\rightarrow$ 63.13k tok/s \\
\bottomrule
\end{tabular}
\end{table}

The training cost is paid once, whereas the serving savings recur on every workflow. In this workload, PrefillShare reduces local prefill from 9.56M to 0.68M tokens per 64-session run and increases throughput from 28.68k to 63.13k tok/s. Under the same GPU budget, the baseline therefore requires roughly twice the GPU time to deliver the same amount of work, allowing the measured 12.1 GPU-hours of cache-conditioned fine-tuning to be recovered within a few hours of sustained serving. This one-time-versus-recurring trade-off is consistent with recent analyses of LLM deployment, which model total deployment cost as a one-time training term plus an inference-volume-weighted recurring term, and with recent work noting that repeated inference can substantially exceed a one-time fine-tuning cost~\citep{mittal2026lifecycle,das2026breakeven}. Thus, the offline fine-tuning cost is amortized as the system serves repeated multi-model workflows rather than recurring on every request.

\begin{table}[t]
\centering
\caption{\textbf{Practical memory accounting for a 4K shared prefix.}
We compare four LLaMA-3.1-8B specialists under the same GPU budget. The baseline stores one shared-prefix KV copy per specialized model, whereas PrefillShare stores one shared copy across models. Resident model-weight memory remains comparable between the two systems.}
\label{tab:memory_accounting}
\small
\setlength{\tabcolsep}{6pt}
\renewcommand{\arraystretch}{1.08}

\begin{tabular}{lcc}
\toprule
\textbf{Memory Component}
& \textbf{Baseline}
& \textbf{PrefillShare} \\
\midrule
Shared-prefix KV copies
& 4
& 1 \\
Prompt KV per session
& 2.0 GiB
& 0.5 GiB \\
Resident model weights
& $\sim$16 GiB / worker
& $\sim$16 GiB / worker \\
\bottomrule
\end{tabular}
\end{table}

\subsection{Practical Memory Accounting}
\label{appendix:memory_accounting}

The main paper reports the asymptotic reduction in KV memory. We additionally account for the absolute KV footprint and resident model weights. For four LLaMA-3.1-8B specialists with a 4K shared prefix, LLaMA-3.1-8B stores 128~KiB of KV state per token. The baseline therefore retains four model-specific copies of the shared prefix, whereas PrefillShare retains a single reusable copy.

The KV-memory reduction is not offset by additional model-weight residency. Under the same 4P+4D GPU budget, the baseline keeps specialized models resident on both prefill and decode workers, whereas PrefillShare keeps the shared base model on prefill workers and task-specific models on decode workers. The resident parameter footprint is therefore comparable between the two systems.

The distinction becomes more important as context length and concurrency increase: model weights remain fixed per worker, whereas aggregate KV memory grows with the number of active sessions and their context lengths. Table~\ref{tab:input_length_ablation} reflects this trend: PrefillShare provides no throughput improvement at 64 input tokens, where prefill is inexpensive, but improves throughput by 2.2$\times$ at 4,096 input tokens. Fig.~\ref{fig:inflight_hit_throughput} similarly shows that avoiding duplicated prefix KV allows PrefillShare to sustain cache reuse over a wider concurrency range.

\subsection{Other Benchmarks and Out-of-Domain Accuracy}
\label{appendix:other_ood}

We additionally evaluate PrefillShare on other math and coding benchmarks and on an out-of-domain task. For math, we use GSM-SEM~\citep{singh2026gsmsem}; for coding, we use CodeEval-Pro~\citep{yu-etal-2025-humaneval}, a harder reconstruction of HumanEval and MBPP. For the out-of-domain evaluation, the Full-FT and PrefillShare models are fine-tuned only on coding data and then evaluated on the non-CS disciplines of SuperGPQA~\citep{pteam2025supergpqascalingllmevaluation}, with no overlap between the fine-tuning and evaluation domains.

\begin{table}[b]
\centering
\caption{\textbf{Accuracy on other benchmarks and an out-of-domain task.}
GSM-SEM evaluates mathematical reasoning, CodeEval-Pro evaluates coding on harder variants of HumanEval and MBPP, and SuperGPQA non-CS evaluates out-of-domain performance after coding-only fine-tuning.}
\label{tab:other_ood}
\small
\setlength{\tabcolsep}{6pt}
\renewcommand{\arraystretch}{1.08}
\begin{tabular}{lccc}
\toprule
\textbf{Method}
& \textbf{GSM-SEM}
& \textbf{CodeEval-Pro}
& \textbf{SuperGPQA non-CS} \\
\midrule
Base
& 61.76
& 52.21
& 24.54 \\
Full-FT
& 76.90
& 59.78
& 24.72 \\
PrefillShare (Ours)
& 76.64
& 59.23
& 24.95 \\
\bottomrule
\end{tabular}
\end{table}

PrefillShare stays within 0.26 and 0.55 points of Full-FT on GSM-SEM and CodeEval-Pro, respectively, extending the accuracy parity observed in the main evaluation to additional math and coding benchmarks. For the out-of-domain evaluation, Base, Full-FT, and PrefillShare remain within 0.41 points on SuperGPQA non-CS. Cache-conditioned fine-tuning therefore preserves the in-domain gains of conventional fine-tuning without introducing additional out-of-domain degradation in this evaluation.

\section{Comparison with Prior KV-Cache Sharing Methods}
\label{appendix:additional_prior_work}

\subsection{Matched Adaptation Comparison with ICaRus}
\label{appendix:icarus_matched}

To isolate the architectural contribution from adaptation capacity, we compare ICaRus and PrefillShare under the same LoRA configuration: rank 128, $\alpha=256$, with LoRA applied to the q, o, gate, up, and down projections. Both methods use the same logical batch size on a single A100 with LLaMA-3.1-8B, TP=1, and input/output lengths of 1024/128 tokens.

\begin{table}[t]
\centering
\caption{\textbf{Matched-adaptation comparison with ICaRus.}
ICaRus and PrefillShare use the same LoRA configuration to control for adaptation capacity. TPOT is measured with input and output lengths of 1,024 and 128 tokens, respectively. Math averages GSM8K and GSM+, coding averages HumanEval and HumanEval+, and tool calling averages the Simple Python and Multiple categories of BFCL.}
\label{tab:icarus_matched}
\small
\setlength{\tabcolsep}{3.5pt}
\renewcommand{\arraystretch}{1.08}
\begin{tabular*}{\linewidth}{@{\extracolsep{\fill}}lcccccc@{}}
\toprule
& \multicolumn{3}{c}{\textbf{TPOT (ms)}}
& \multicolumn{3}{c}{\textbf{Accuracy}} \\
\cmidrule(lr){2-4}
\cmidrule(lr){5-7}
\textbf{Method}
& \textbf{B=1}
& \textbf{B=4}
& \textbf{B=16}
& \textbf{Math}
& \textbf{Coding}
& \textbf{Tool} \\
\midrule
ICaRus (LoRA)
& 16.5 & 19.9 & 22.9
& 59.1 & 46.1 & 88.3 \\
PrefillShare (matched LoRA)
& 15.8 & 19.6 & 20.7
& 60.0 & 45.7 & 89.0 \\
PrefillShare (Full-FT)
& 12.1 & 15.5 & 14.8
& 60.4 & 47.0 & 89.6 \\
\bottomrule
\end{tabular*}
\end{table}

At equal adaptation budget, PrefillShare has lower TPOT at every evaluated batch size, including about 10\% at B=16, while maintaining comparable task accuracy (+0.9 Math, +0.7 Tool, $-0.4$ Coding). This matched comparison removes adaptation capacity as the explanation for the serving gap. The remaining difference is architectural. ICaRus keeps the frozen base model on the decode path, so every step runs both the base model for compatible KV and the base-plus-adapter path for generation. PrefillShare separates the phases completely: the base model runs only during prefill and the task-specific model runs during decode, so generation carries no additional base-model computation.

The separation also removes the adapter-based specialization constraint. ICaRus requires the base-model path to remain compatible with the task-specific path during generation, whereas PrefillShare can use full fine-tuning for the task-specific decode model. The Full-FT row therefore represents a configuration that PrefillShare can support without retaining an additional base-model path during decode.

\subsection{Comparison with KVCOMM and DroidSpeak}
\label{appendix:kvcomm_droidspeak}

We further compare PrefillShare with other KV-cache sharing algorithms,
including KVCOMM~\citep{ye2025kvcomm} and DroidSpeak~\citep{liu2024droidspeak}.
A direct end-to-end quantitative comparison is difficult because the methods
differ in target setting and implementation availability. KVCOMM focuses on
anchor-based KV reuse across different system prompts, while PrefillShare targets
shared-prefix reuse across differently fine-tuned task-specific decoders in
multi-model agent workloads. Applying KVCOMM to this setting would require
adapting its anchor-based reuse mechanism, making the result sensitive to
nontrivial design choices. DroidSpeak is more closely related in target setting,
but its official implementation is not publicly available, preventing a faithful
reproduction. We therefore provide a normalized qualitative and asymptotic
comparison in Table~\ref{tab:kv_reuse_comparison}.

\begin{table*}[t]
\centering
\renewcommand{\arraystretch}{1.15}
\setlength{\tabcolsep}{5pt}
\caption{\textbf{Comparison of KV reuse algorithms in a multi-agent system.}
$N$ and $L$ denote the number of agents and sequence length, respectively; $V$ and $r$ denote the number of anchors and anchor retrieval hit rate in KVCOMM; and $\alpha$ denotes the fraction of sensitive layers in DroidSpeak.}
\label{tab:kv_reuse_comparison}
\small
\begin{tabular}{
p{0.17\textwidth}
p{0.25\textwidth}
p{0.25\textwidth}
p{0.25\textwidth}}
\toprule
\textbf{Aspect}
& \textbf{KVCOMM}
& \textbf{DroidSpeak}
& \textbf{PrefillShare (Ours)} \\
\midrule
Method
& Anchor-based KV reuse
& Selective recomputation
& KV-sharing-aware fine-tuning \\

Target
& Different system prompts
& Different fine-tuned models
& Different fine-tuned models \\

Training Overhead
& Training-free
& Training-free
& Fine-tuning \\

Inference Overhead
& Anchor lookup
& Calibration
& No overhead \\

KV sharing
& Not full KV sharing (anchor-guided KV reuse)
& Not full KV sharing (layer-wise KV reuse)
& Full KV sharing \\

Accuracy
& Limited by train-test discrepancy
& Limited by train-test discrepancy
& No train-test discrepancy; higher accuracy \\

KV memory
& $O((1+V)L)$
& $O((1+\alpha(N-1))L)$
& $O(L)$ \\

\# Prefill Compute
& $O((1+(1-r)(N-1))L^2)$
& $O((1+\alpha(N-1))L^2)$
& $O(L^2)$ \\
\bottomrule
\end{tabular}
\end{table*}

Table~\ref{tab:kv_reuse_comparison} compares the methods under a common multi-agent abstraction with $N$ agents and sequence length $L$. KVCOMM and DroidSpeak have the practical advantage of being training-free. However, their reuse is conditional or partial: KVCOMM depends on anchor retrieval, and DroidSpeak recomputes sensitive layers. These inference-time costs are paid repeatedly during serving. PrefillShare makes a different trade-off by introducing cache-conditioned fine-tuning, an offline cost that is amortized after deployment. This training aligns the decoder with the shared base KV cache, enabling full shared-prefix KV reuse without additional anchor lookup or layer-wise recomputation during inference.

Under this abstraction, KVCOMM requires memory proportional to the anchor pool,
yielding $O((1+V)L)$ KV memory, and its prefill compute depends on the anchor
hit rate $r$, giving $O((1+(1-r)(N-1))L^2)$. DroidSpeak reduces redundant
prefill by reusing non-sensitive layers, but still recomputes an $\alpha$
fraction of sensitive layers for each additional agent, resulting in
$O((1+\alpha(N-1))L)$ KV memory and
$O((1+\alpha(N-1))L^2)$ prefill compute. PrefillShare stores the shared-prefix
KV cache once and performs the shared prefill once, leading to $O(L)$ KV memory
and $O(L^2)$ prefill compute for the shared context. Thus, while training-free
methods are attractive for deployment simplicity, PrefillShare shifts the cost
to offline fine-tuning and reduces repeated inference-time overhead in
multi-agent shared-context serving.
%%%%%%%%%%%%%%%%%%%%%%%%%%%%%%%%%%%%%%%%%%%%%%%%%%%%%%%%%%%%%%%%%%%%%%%%%%%%%%%
%%%%%%%%%%%%%%%%%%%%%%%%%%%%%%%%%%%%%%%%%%%%%%%%%%%%%%%%%%%%%%%%%%%%%%%%%%%%%%%

\end{document}